\documentclass[conference]{IEEEtran}
\IEEEoverridecommandlockouts
\usepackage{times}
\usepackage{graphicx} 
\usepackage{epstopdf}
\usepackage{subfigure}
\usepackage{color}

\usepackage[ruled]{algorithm2e} 
\usepackage{array}
\usepackage{amsmath}
\usepackage{bm}
\usepackage{bbm}
\usepackage{amsfonts}
\usepackage{amssymb}
\usepackage{multirow}
\usepackage{multicol}
\usepackage{booktabs}

\newcommand{\fnorm}[1]{\ensuremath{\|#1\|_F}}

\begin{document}

\title{Supervised Nonnegative Matrix Factorization to Predict ICU Mortality Risk\\
}

\author{\IEEEauthorblockN{ Guoqing Chao}
\IEEEauthorblockA{\textit{Feinberg School of Medicine} \\
\textit{Northwestern University}\\
Chicago, U.S \\
guoqingchao10@gmail.com}
\and
\IEEEauthorblockN{ Chengsheng Mao}
\IEEEauthorblockA{\textit{Feinberg School of Medicine} \\
\textit{Northwestern University}\\
Chicago, U.S \\
chengsheng.mao@northwestern.edu}
\and
\IEEEauthorblockN{ Fei Wang}
\IEEEauthorblockA{\textit{Weill Cornell Medicine} \\
\textit{Cornell University}\\
New York, U.S \\
few2001@med.cornell.edu}
\and
\IEEEauthorblockN{ Yuan Zhao}
\IEEEauthorblockA{\textit{Feinberg School of Medicine} \\
\textit{Northwestern University}\\
Chicago, U.S \\
yuan.zhao1@northwestern.edu}
\and
\IEEEauthorblockN{ Yuan Luo}
\IEEEauthorblockA{\textit{Feinberg School of Medicine} \\
\textit{Northwestern University}\\
Chicago, U.S \\
yuan.luo@northwestern.edu}

}

\maketitle

\begin{abstract}
ICU mortality risk prediction is a tough yet important task. On one hand, due to the complex temporal data collected, it is difficult to identify the effective features and interpret them easily; on the other hand, good prediction can help clinicians take timely actions to prevent the mortality. These correspond to the interpretability and accuracy problems. Most existing methods lack of the interpretability, but recently Subgraph Augmented Nonnegative Matrix Factorization (SANMF) has been successfully applied to time series data to provide a path to interpret the features well. Therefore, we adopted this approach as the backbone to analyze the patient data. One limitation of the original SANMF method is its poor prediction ability due to its unsupervised nature. To deal with this problem, we proposed a supervised SANMF algorithm by integrating the logistic regression loss function into the NMF framework and solved it with an alternating optimization procedure. We used the simulation data to verify the effectiveness of this method, and then we applied it to ICU mortality risk prediction and demonstrated its superiority over other conventional supervised NMF methods. 

\end{abstract}


\section{Introduction}

With the fast development of machine learning and data mining, automated predictive modeling becomes possible when combining with the increasingly available medical data in hospitals and clinical institutions. These models can discover latent patterns hidden in the data and help clinicians make timely and accurate decisions. The data generated in the Intensive Care Unit (ICU) are abundant and complicated due to the diverse variable types (especially the continuous time series data). One key event in ICU is patient death. However, the complex nature of the ICU data usually makes the clinicians difficult to make timely and correct decisions. This makes predictive modeling of the mortality risk an important problem. 

In this paper, we describe a supervised nonnegative matrix factorization (NMF) algorithm that performs predictive modeling by the exploration of atomic and the higher-order features jointly. We automate the mining of higher-order features by converting data into graph representation. Moreover, supervised latent group identification reduces dimensionality of different feature types for patients, and simultaneously group temporal trends to form effective features for patient outcome prediction. Applications on patient physiological time series~\cite{luoaai2016} show significant performance improvements from multiple baselines.

\section{Related work}

Nonnegative Matrix Factorization (NMF) refers to the set of problems on approximating a nonnegative matrix as the product of several nonnegative matrices. The problem has become popular since the work ~\cite{lee1999nature}, where they form a nonnegative matrix by concatenating the set of pixel intensity vectors stretched from human facial images. After factorizing such matrix into the product of two matrices, they found that one matrix can be interpreted as the set of image basis with part based representation of human faces, and the other matrix has the coefficients if we were to reconstruct the face image from those bases. Because of the nonnegativity constraints, NMF is not a convex problem and they developed a multiplicative updates algorithm to obtain a stationary solution, and they prove the convergence of the algorithm in their paper~\cite{lee2001nips}.

Since then people have been working on NMF from various aspects. ~\cite{ding2005sdm} showed that there is some equivalence between NMF and K-means/spectral clustering and thus claimed NMF can be used for data clustering purpose~\cite{chao2018discriminative}. ~\cite{ding2006kdd} further developed a t-NMF approach that can perform co-clustering on both matrix columns and rows. They also discussed the various variants of NMF in~\cite{ding2010pami}. ~\cite{dhillon2005nips} extended NMF to the case when the matrix approximation loss is measured by Bregman divergence, which is a much more general loss with both Frobenius norm and KL divergence that was introduced in ~\cite{lee2001nips} as its special cases. On the solution procedure aspect, multiplicative updates has been recognized for its slow convergence and poor quality. ~\cite{berry2007algorithms} reviewed the general algorithms before 2007, three classes of algorithms are categorized. The first class is multiplicative updates, the second class is gradient based methods such as ~\cite{hoyer2004non,chu2004optimality,pauca2006nonnegative}, the third class is the alternating least squares (ALS) algorithm~\cite{paatero1994positive,paatero1999multilinear,langville2014algorithms}. Also in 2007,~\cite{lin2007nc} proposed an efficient projected gradient approach for NMF, which adopted Taylor expansion strategy to reduce the cost. ~\cite{kim2011sdm} also proposed an active set type of method called principal block pivoting to solve the NMF problem. ~\cite{sun2014icassp} adopted the alternating direction method of multipliers (ADMM) to solve the NMF with beta-divergence.  Hsieh and Dhillon~\cite{hsieh2011kdd} designed a fast coordinate descent method to accelerate the FastHals~\cite{cichochi2009ieice}. 

Apart from basic NMF method introduced above, there are also many variants. They can be grouped into three groups. The first group enforced some constraints into basic NMF to obtain certain desirable charateristics, such as sparse NMF~\cite{hoyer2004non,hoyer2002non,morup2008approximate}, orthogonal NMF~\cite{li2010nonnegative,ding2006orthogonal,yoo2010orthogonal}, discrininant NMF~\cite{wang2004accv,zafeiriou2006TNN} and manifold NMF~\cite{cai2008non,cai2011graph}. The second group named structured NMF modified the standard formulation of NMF, including weighted NMF~\cite{kim2009weighted}, convolutive NMF~\cite{o2006convolutive} and nonnegative matrix trifactorization~\cite{yoo2010orthogonal}. The third group is the generalized NMF, including semi-NMF~\cite{ding2010convex}, nonnegative tensor factorization~\cite{cichocki2008nonnegative}, matrix-set factorization~\cite{li2007non} and kernel NMF~\cite{zhang2006non}. For details, refer to the NMF survey paper~\cite{wang2013nonnegative}.

Although these general NMFs decrease the dimension of the original data successfully, it cannot guarantee that the prediction ability is retained because they're unsupervised methods that have not used the label information. To tackle this problem, several supervised NMF methods were proposed. They can be classified into two categories. The first category included~\cite{wang2004accv,zafeiriou2006TNN,xue2006icpr} is the method incorporating the idea of linear discriminative analysis to improve the prediction ability of the reduced representation, while the second category included ~\cite{lee2010spi,jing2012tip} is the method introducing a loss function to take the label information into consideration.

Our proposed supervised NMF belongs to the second category. The previous two methods ~\cite{lee2010spi,jing2012tip} enforced a frebenius loss to constrain the label information, however, it seems more like the regression not classification. Therefore, the frebenius loss constraint is not enough to exploit the label information for classification. Instead of the frebenius loss constraint, we enforce the classification loss constraint explicitly to guarantee the prediction ability of the learned representation. In fact, we proposed a supervised NMF framework that can incorporate any classification loss function. In this paper, we implement one with the logistic regression, and other classification loss functions can be considered to be integrated in furture work.

\section{Methods}

\subsection{NMF and Logistic Regression}
NMF~\cite{lee1999nature} is a popular representation learning technique. Assume that we have data matrix $\bm X\in \mathcal{R}^{n\times m}$ and its corresponding label $y \in \mathcal{R}^n$. NMF aims to learn two nonnegative matrices $\bm U\in \mathcal{R}^{n\times r}$  and $\bm V\in \mathcal{R}^{r\times m}$ to approximate a nonnegative matrix $\bm X\in \mathcal{R}^{n\times m}$, each row of which  contains $m$ features of one of the $n$ samples. Then NMF can be formulated in matrix form $\bm X = \bm U\bm V$, and at the same time $\bm U$ and $\bm V$ are constrained to be nonnegative. Therefore, NMF is formulated as the following optimization problem. 
\begin{equation}
\label{nmf}
\begin{split}
&\underset{\bm U,\bm V}{\mbox{min}}\fnorm{\bm X - \bm U\bm V}^2\\
& s.t. \bm U\succeq  0, \bm V\succeq  0,
\end{split}
\end{equation}
where $\fnorm{\cdot}^2$ indicates squared frobenius norm and $\bm U\succeq  0, \bm V\succeq  0$ mean that $\bm U$ and $\bm V$ are both entry-wise nonnegative. $r$ rows of $\bm V$ are considered as new bases while $r$ columns of $\bm U$ are viewed as the coefficients with which the original samples can be represented with a linear combination of bases. $\bm U$ can be considered as a low-dimensional representation of $\bm X$ since generally $r< m$. 

It can be seen that NMF cannot utilize the label information to learn the representation and thus it cannot guarantee the classification performance. To take the advantage of the supervised information, we will introduce the supervised learning methods, precisely, the loss function for supervised learning methods. Supervised learning is the learning paradigm that uses labels to learn, like classification. Herein, we will introduce a popular supervised learning method Logistic regression (LR).

Different from NMF, LR  can make good use of the label information to classify. The optimization problem of LR can be formulated as follows:
\begin{equation}
\label{lrobj}
\begin{split}
\underset{\bm w,b}{\mbox{min}}\sum_{i=1}^n \mbox{ln}\Big(1+\mbox{exp}(-y_i\big(\sum_{j=1}^r w_j \bm X_{ij}+b)\big)\Big) + \frac{1}{2}\beta (\sum_{j=1}^r w_j^2+b^2)
\end{split}
\end{equation}
where $\bm X_{i, j}$ indicates the $j$th feature of $i$th data point of the $\bm X$, $\bm w$ and $b$ indicate the  weight and bias of LR. In Eq.~\eqref{lrobj} the first item is the loss function while the second one is the regularization item to prevent over-fitting. Parameter $\beta$ is used to balance the loss and regularization items.   
\subsection{Supervised NMF Problem Formulation}
Given data matrix $X^{n\times m}$ and label vector $y^{n\times 1}$, where $n$ is the number of patients, $m$ is the number of subgraphs, and the number of the labels is 1. Note that entries of $y$ is 1 or 0.
Coefficient matrix $U^{n\times r}$, basis matrix $V^{r\times m}$ and weight matrix $w^{r\times 1}$, where $r$ is the number of subgraph groups.
The supervised nonnegative matrix factorization (SNMF) can be formulated as
\begin{equation}
\label{obj}
\begin{split}
&\underset{\bm U,\bm V,\bm w,b}{\mbox{min}} L_f + L_{lr} + L_r\\
&s.t. \bm U\succeq  0, \bm V\succeq  0,
\end{split}
\end{equation}
where the first item $L_f = \frac{1}{2}\fnorm{\bm X-\bm U\bm V}^2$ is the loss function for NMF, the second item $L_f = \alpha\sum_{i=1}^n \mbox{ln}\Big(1+\mbox{exp}(-y_i\big(\sum_{j=1}^r w_ju_{ij}+b)\big)\Big)$  is the loss function for LR and the third item $L_r = \frac{1}{2}\beta (\sum_{j=1}^r w_j^2+b^2)+\frac{1}{2}\gamma \fnorm{\bm U}^2$ is the regularization for NMF and LR.

Obviously, NMF and LR are integrated into this united framework. Since $\bm U$ is the new representation we aim to learn, LR works on it instead of original data matrix $\bm X$. The last item is used to regularize $\bm U$. $\alpha$, $\beta$ and $\gamma$ are used to balance the role of those corresponding items.
\subsection{Optimization}
For the supervised NMF optimization problem~\eqref{obj}, we can find that NMF and LR objectives are integrated together. In introduction part we have mentioned that projected gradient descent method~\cite{lee2010spi} can solve this problem and gradient descent method is also the general algorithm to solve LR. we can also see that some of the variables $\bm U$, $\bm V$, $\bm w$ and $b$ are interwined with each other, alternative minimization is an suitable tool to solve this kind of problems.

Our optimization will be split into four subproblem minimizations: minimize $\bm U$ with fixed $\bm V$, $\bm w$, and $b$; minimize $\bm V$ with fixed $\bm U$, $\bm w$, and $b$; minimize $\bm w$ with fixed $\bm U$, $\bm V$, and $b$; minimize $b$ with fixed $\bm U$, $\bm V$, $\bm w$. These four subproblem minimizations will be alternatively executed until the predefined termination condition is satisfied. For each of the four steps, projected gradient descent or gradient descent will be adopted to solve the subproblems. 

The gradients of the objective function in Eq.~\eqref{obj} with respect to the four variables are given below, and the detailed derivations are provided in Appendix.
\begin{equation}
\begin{split}
\nabla_{\bm U}L=-\bm X{\bm V}^T+\bm U\bm V{\bm V}^T+\gamma \bm U-\alpha (\bm y {\bm w}^T)\oslash\bm {D}.
\end{split}
\end{equation}

\begin{equation}
\begin{split}
\nabla_{\bm V}L=-{\bm U}^T\bm X+{\bm U}^T\bm U\bm V.
\end{split}
\end{equation}

\begin{equation}
\begin{split}
\nabla_{\bm w}L=-\alpha(\bm U\odot\bm Y\oslash \bm {D})^T* \bm {e_n}+\beta \bm w.
\end{split}
\end{equation}

\begin{equation}
\begin{split}
\nabla_{\bm b}L=-\alpha \bm {e_n}^T*\Big(\bm y\oslash \big(1+\mbox{exp}(\bm U\bm w+b)\odot \bm y\big)\Big)+\beta b.
\end{split}
\end{equation}
where $\oslash$ indicates element-wise division and $\odot$ indicates element-wise multiplication. $\bm {D}$ is obtained by repeating the column vector $(1+\mbox{exp}{(\bm U\bm w+b \bm {e_n}))\odot \bm y}$ $r$ times to form the matrix of size $n\times r$. $\bm Y$ is obtained with the same operator to vector $\bm y$.

The update formulation of these four variables are given below:
\begin{equation}
\label{Uupdateformulation}
\begin{split}
{\bm U}^{t+1}=\mathcal{P}_{+}[{\bm U}^t-\eta_{\bm U}^t\nabla_{\bm U}L({\bm U}^t,{\bm V}^t, \bm w^t, b^t)]
\end{split}
\end{equation}

\begin{equation}
\label{Vupdateformulation}
\begin{split}
{\bm V}^{t+1}=\mathcal{P}_{+}[{\bm V}^t-\eta_{\bm V}^t\nabla_{\bm V}L({\bm U}^{t+1},{\bm V}^t, \bm w^t, b^t)]
\end{split}
\end{equation}

\begin{equation}
\label{wupdateformulation}
\begin{split}
\bm w^{t+1}= \bm w^t-\eta_{\bm w}^t\nabla_{\bm w}L({\bm U}^{t+1},{\bm V}^{t+1}, \bm w^t, b^t)
\end{split}
\end{equation}

\begin{equation}
\label{bupdateformulation}
\begin{split}
b^{t+1}= b^t-\eta_b^t\nabla_bL({\bm U}^{t+1},{\bm V}^{t+1}, \bm w^{t+1}, b^t)
\end{split}
\end{equation}

The scalar form of the problem Eq.\eqref{obj} can be written in matrix form as follows.
\begin{equation}
\label{objmatrix}
\begin{split}
&\underset{\bm U,\bm V,\bm w,b}{\mbox{min}} L_f + L_{lr}+L_r\\
&s.t. \bm U\succeq  0, \bm V\succeq  0,
\end{split}
\end{equation}
where the first item  $L_f=\frac{1}{2}\fnorm{\bm X-\bm U\bm V}^2$ indicates the frobenius loss function for NMF, the second item $L_{lr}=\alpha {\bm {e_n}}^T*\mbox{ln}\Big(1+\mbox{exp}(-(\bm U\bm w+b\bm {e_n})\odot \bm y)\big)\Big)$ indicates the loss function for LR and the third item $L_r=\frac{1}{2}\beta ({\bm w}^T\bm w+b^2)+\frac{1}{2}\gamma \fnorm{\bm U}^2$ indicates the regulairzation for NMF and LR.

To understand the algorithm clear, we summarized the algorithm in Algorithm 1. 
        \begin{algorithm}
            \caption{Supervised Non-Negative Matrix Factorization}
            \KwIn{$\bm X, \bm Y, \alpha, \beta, \gamma, \lambda $}
            \KwOut{$\bm U, \bm V$}
            
            {\bf Initialization}: Initialize $\bm U$ and $\bm V$ as $\bm U_0$ and $\bm V_0$ with the algorithm NNDSVD and initialize $w$ and $b$ as $w_0$ and $b_0$\;
            \While{not reach the maximal step}{
				calculate $n$: the sum of the frobenius norms of $\bm U$, $\bm V$, $w$ and $b$.
                \eIf{n is not less than the tolerance}
                {
                1. Update ${\bm U}^t$ according to Eq.~\eqref{Uupdateformulation}\;
                2. Update ${\bm V}^t$ according to Eq.~\eqref{Vupdateformulation}\;
                3. Update $w^t$ according to Eq.~\eqref{wupdateformulation}\;
                4. Update $b^t$ according to Eq.~\eqref{bupdateformulation}\;
                }
                {
                    break\;
                }
        }
    \end{algorithm}
\section{Experiments}
Before applying the proposed method SNMF to the ICU mortality risk prediction problem, we verify the effectiveness of SNMF on the simulation data.
\subsection{Simulation study}

To demonstrate the effectiveness of the proposed supervised NMF method, we generate a simulation data to verify it. The data generating process includes three steps: Firstly,
the coefficient matrix $\bm U$ is generated with two Gaussian distributions to indicate  the true latent representation, the mean vectors and covariance matrices of the two Gaussian distributions are $\bm \mu_1 = (1, 1, 1, 1, 1)$, $ \bm {\Sigma1} = \bm {\mbox{diag}}(1,1,1,1,1)$ and $\bm \mu_2 = (3, 3, 1, 1, 1)$, $ \bm {\Sigma2} = \bm {\mbox{diag}}(1,1,1,1,1)$, respectively. 250 points are generated for each group, so that $\bm U_1\in \mathcal{R}^{250\times5}$ and $\bm U_2\in \mathcal{R}^{250\times5}$ indicate the coefficient matrices corresponding to two groups of data. In order to guarantee $\bm U$  nonnegative, all the negative entries are set to zero. Here, we use vector $\bm y$ to indicate which Gaussian distribution the point is from and form the labels of the generated data.  Secondly, the basis matrix $\bm V\in \mathcal{R}^{5\times10}$ is generated with the Uniform distribution in $(0, 1)$ and set all the negative entries to zero to make sure $\bm V$ is nonnegative. Thirdly, the noise $\bm e$ is generated using a Gaussian distribution with the same mean and covariance matrix with $\bm U*\bm V$ and then added into the matrix $\bm U*\bm V$, see Eq.~\eqref{noiseadding}.

\begin{equation}
\label{noiseadding}
\begin{split}
\bm X = \bm U*\bm V+\eta* e
\end{split}
\end{equation}
where $e$ is the noise from the Gaussian distribution $\mathcal{N}(\bm \mu,\bm \Sigma)$ and $\bm \mu$ and $\bm \Sigma$ indicate the mean and covariance matrix of data matrix $\bm U*\bm V$, $\eta$ indicates the noise level. Final data $\bm X$ is $\mathcal{R}^{500\times10}$. It should be noted that all the negative entries of $\bm X$ are set to zero to obtain the nonnegative matrix $\bm X$. $\bm y$ takes values 0 or 1 to indicate the label. If the data point is generated from the first Gaussian distribution, its label will be 0 otherwise 1.

We split the obtained data matrix $\bm X$ and corresponding labels $\bm y$ into training and test sets with the ratio half to half, and then we compared the representation obtained from unsupervised NMF with our proposed SNMF. In addition, another supervised NMF method FNMF~\cite{zafeiriou2006TNN} is also added to compare. For the generated coefficient matrix  $\bm U$, we know there are 2 distinguished dimensions, to make sure these methods can capture this distinguished information,  for all the four methods, we set the dimension of the new representation as 2. For our proposed SNMF, the range of parameters $\alpha,\beta, \gamma$ are from $\{0, 0.001, 0.01, 0.1\}$. It should be noted that when all of $\alpha,\beta, \gamma$ take value 0, SNMF degenerates to NMF, and in our experimental results we adopt this as unsupervised NMF. To tune parameters efficiently, the parameters $\alpha, \beta, \gamma$ are set to $\alpha/p$, $\beta/r$ and $\gamma/(n\times r)$ where $p$ is the number of the features of data matrix $\bm X$, $r$ is the number of the features of the latent representation matrix $\bm U$ and $n$ is the number of the samples of data matrix $\bm X$. To further explore how these methods perform on the data with different noise level, we conduct the experimental comparison on the simulation data with three noise level 0, 0.2, and 0.5. The metric we adopt is the area under ROC (receiver operating characteristic) curve (AUC). All the experimental results are demonstrated in Table~\ref{table_summary}, the best result is shown in bold font for each case.

\begin{table}[!htbp]
\caption{The AUC on the simulation data ($\%$), the value in bold indicates the difference with the rest is significant ($p$ value $<$0.01).}\label{table_summary}
\centering
\begin{tabular}{|c|c|c|c|c|c|c|c|}
  \hline
  Noise Level & Data Split & NMF & FNMF  & SNMF\\
  \hline
    $\eta =0.5$ & Test& 90.35 & 89.19 & \textbf{91.78} \\
  \hline
      $\eta =0.2$ & Test  & 93.91 & 94.40 & \textbf{95.68} \\
  \hline
      $\eta =0$ & Test  & 95.23 & 95.96 & \textbf{97.15}\\
  \hline
\end{tabular}
\end{table}

From Table~\ref{table_summary}, we can find that on the data without any noise, FNMF performs better than NMF but worse than our SNMF, because FNMF learned the new representation with the help of the label information while NMF doesn't. Compared with SNMF, FNMF just guarantees the data points within the same class close and those in different classes far away in the new representation but it still doesn't use the label information directly to approach the final goal classification, so SNMF is more effective. With increasing the noise level, we can see all the methods' performance degrade, but SNMF outperforms all the other methods all the time. During all the comparisons, NMF performs worse than SNMF is because when it learned the 2 dimensional new representation, it maybe not exploit the distinguished dimensional information sufficiently because of its unsupervised property, but for SNMF, it makes full use of that distinguished dimensional information due to the classification loss item in the Eq.~\eqref{obj}. 

\subsection{ICU Mortality Risk Prediction}
\subsubsection{Data Processing}
\begin{figure}[!htbp]
\begin{center}
\centerline{\includegraphics[width=0.9\columnwidth,height=0.5\columnwidth]{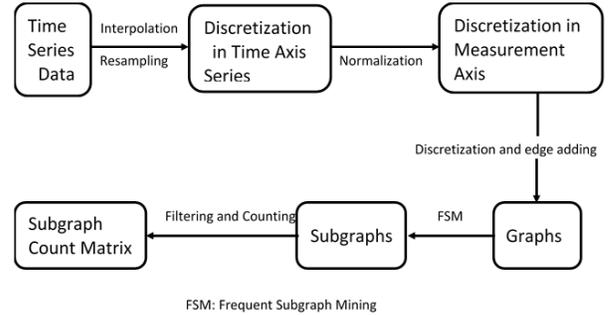}}
\centering
\caption{Schematic of the time series data processing.}
\label{fig:schematic}
\end{center}
\end{figure}

The data we adopted to predict ICU mortality risk is processed as in Paper~\cite{luoaai2016}. To make the data processing clearly, a schematic of the data processing workflow is illustrated in Figure~\ref{fig:schematic}. The time series data from the second half of the first day after patients' admissions to ICU are the original data we need to process. To address the issues of missing data and irregular sampling interval, we performed linear interpolation and resampling at regularly spaced time intervals on the original time series to discretize in the time axis. Although a more complex  imputation algorithm may be more plausible~\cite{luo2016using,luo20173d}, we follow the SANMF approach to enable fair comparison. Note that the time interval 2 hours are determined by 5-fold cross validation over choices of 1, 2, 4, or 6 hour intervals. For detailed processing information, the authors can refer to Paper~\cite{luoaai2016}.   
%
%

\subsubsection{Experiment Setting and Result}
\begin{figure} [!htbp]
\begin{center}
\centerline{\includegraphics[height =2.5in, width=3.5in]{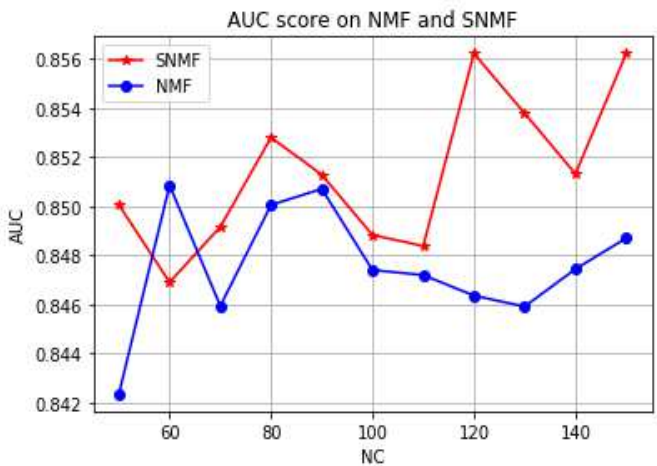}}
\caption{Average AUC comparison under different number of groups on the hold-out test data.} \label{fig:subfig} 
\end{center}
\end{figure}

The processed subgraph count data are split equally, stratified by mortality, into the training set and test set. There are 3932 cases in the training set while 3931 cases in the test set. The statistics of the data are demonstrated in Table~\ref{tab:statistics}. For our proposed SNMF, we need to learn the new representation $\bm U$ from the training data and then train the logistic regression on top of $\bm U$ to check its performance on the $\bm U$ obtained on test data. Besides parameter $r$ (the number of groups), our SNMF still have parameters $\alpha,\beta, \gamma$, we will use 5-fold cross validation to identify the parameters $\alpha,\beta, \gamma$. The range of parameters $\alpha,\beta, \gamma$ are from $-2$ to $2$ with exponential base 10. To tune these parameters efficiently, the parameters $\alpha, \beta, \gamma$ are set to $\alpha/p$, $\beta/r$ and $\gamma/(n\times r)$, which is the sampe with that in simulation data. For the number of groups, we will follow the way in~\cite{luoaai2016}, and extend its range if the performance will still rise up to 120. From the subsquent Figure~\ref{fig:subfig}, we can find that the range for number of groups are from 50 to 150 (We did not show the performance in range 10 to 40 because their performance is worse than the current range). The metric we adopt is the area under ROC (receiver operating characteristic) curve (AUC). All the experimental results are demonstrated in Figure~\ref{fig:subfig}.

From Figure~\ref{fig:subfig}, we can see that almost with each group number, our proposed SNMF outperforms NMF, this should because SNMF take the label information into consideration when learning the low-dimensional representation while NMF doesn't. In some case NMF can perform a bit better than SNMF on the hold-out test data, because there are additional 3 parameters to tune for SNMF and the distributions between training and test data may be inconsistent and this may cause the parameters identified during cross validation not perform well on the hold-out test data. Figure~\ref{fig:subfig} shows that the best test AUC 0.8562 for SNMF occurs at 120 groups while that best test AUC 0.8508 for NMF happens at 60 groups, and the superiority of SNMF over NMF on ICU data is statistically significantly (p-value of permutation test is 0.0307).

\begin{table}[htp]
\caption{Statistics of Experimental Data}
\begin{center}
\begin{tabular}{|c|c|c|c|}
\hline
Mortality & Total Cases & Training Cases & Test Cases\\
\hline
$\leq$ 30 days & 788 & 383(9.7$\%$) & 405(10.3 $\%$)\\
\hline
$>$ 30 days or alive & 7075 & 3549(90.3$\%$) & 3526(89.7 $\%$)\\
\hline
\end{tabular}
\end{center}
\label{tab:statistics}
\end{table}%

\section{Discussion} 
From the classification results on the ICU data, our SNMF indeed performs better than the unsupervised NMF counterpart from the statistic significancy perspective. However, the benifit is not big, this maybe because the information NMF retained has already possessed discriminative classification prediction ability. 

As for the ICU data, its original data form is time series, the processing workflow proposed in paper~\cite{luoaai2016,xue2018predicting} indeed opened a feasible way to deal with the prediction problem easily, meanwhile, some discriminative information may be lost during the process, such as the interpolation and sampling phases may be not precise. A direct way to deal with time series data and conduct classification is interesting and promising. In addition, feature selection~\cite{dai2015feature,hu2016feature} is closely related to the present work, it can be further compared with the proposed method while working on time series data directly in future work.

There are several different data types of ICU data, like vital sign and lab test information, they can be considered as multi-view information, and then multi-view learning methods~\cite{chao2016consensus,chao2016multi,luo2017tensor,luo2016tensor} can be considered to be adopted to solve this problem. 
\section{Conclusions}
In this paper, we have proposed a supervised NMF method to learn a discriminative representation, based on which the classification performance is improved compared with its unsupervised NMF counterpart and other supervised NMF method. We adopted the projected gradient descent algorithm to solve this problem. The results on synthetic data and ICU data verified its superiority. The learned representation with our method has better prediction ability, which can guide the clinician well in reality.

Since we just explored to integrate NMF with the simple classifier logistic regresssion, other advanced classifier (classification loss function) can be used to replace logistic regression. For instrance, as with the fast development of deep learning, it is promising to combine deep neural network with NMF to learn a more discriminative representation to boost the prediction performation of the ICU mortality risk further. 
\bibliographystyle{IEEEtran} 
\bibliography{snmf}

\appendix
For convenience to derive the gradients with respect to each variable, let $L_X=\frac{1}{2}\fnorm{\bm X-\bm U\bm V}^2+\frac{1}{2}\gamma \fnorm{\bm U}^2$ and $L_y=\alpha\sum_{i=1}^n \mbox{ln}\Big(1+\mbox{exp}(-y_i\big(\sum_{j=1}^r w_ju_{ij}+b)\big)\Big)+\frac{1}{2}\beta (\sum_{j=1}^r w_j^2+b^2)$, $L=L_X+L_y$ denotes the objective function in Eq~\eqref{objmatrix}
.
\begin{equation}
\label{gradUx}
\begin{split}
&\nabla_{\bm U}L_X=\frac{1}{2}\nabla_U \big(\mbox{Tr}\big((\bm X-\bm U\bm V)(\bm X-\bm U\bm V)^T\big)+\gamma \mbox{Tr}(\bm U {\bm U}^T)\big)\\
&=\frac{1}{2}\nabla_{\bm U}\big(\mbox{Tr}(-\bm X{\bm V}^T-\bm U\bm V\bm X+\bm U\bm V {\bm V}^T{\bm U}^T)+\gamma \mbox{Tr}(\bm U {\bm U}^T)\big)\\
&=\frac{1}{2}(-2\bm X{\bm V}^T+2\bm U\bm V{\bm V}^T+2\gamma \bm U)\\
&=-\bm X{\bm V}^T+\bm U\bm V{\bm V}^T+\gamma \bm U
\end{split}
\end{equation}

\begin{equation}
\label{gradV}
\begin{split}
&\nabla_{\bm V}L_X=\frac{1}{2}\nabla_V \Big(\mbox{Tr}\big((\bm X-\bm U\bm V)(\bm X-\bm U\bm V)^T\big))\Big)\\
&=\frac{1}{2}\nabla_{\bm V}\Big(\mbox{Tr}(-\bm X{\bm V}^T-\bm U\bm V{\bm X}^t+\bm U\bm V {\bm V}^T{\bm U}^T)\Big)\\
&=\frac{1}{2}(-2{\bm u}^T\bm X+2{\bm U}^T\bm U\bm V)\\
&=-{\bm U}^T\bm X+{\bm U}^T\bm U\bm V
\end{split}
\end{equation}

\begin{equation}
\label{gradW}
\begin{split}
&\nabla_{w_j}L_y=\alpha \sum_{i=1}^n\frac{\mbox{exp}\big(-y_i(\sum_{j=1}^r w_ju_{ij}+b)\big)*(-1)*y_iu_{ij}}{1+\mbox{exp}\big(-y_i(\sum_{j=1}^r w_ju_{ij}+b)\big)}\\
& \quad \quad  \quad \quad +\beta w_j\\
& \quad \quad  \quad =-\alpha\sum_{i=1}^n\frac{y_iu_{ij}}{1+\mbox{exp}\big(y_i(\sum_{j=1}^r w_ju_{ij}+b)\big)}+\beta w_j
\end{split}
\end{equation}

\begin{equation}
\label{gradWmatrix}
\begin{split}
\nabla_{\bm w}L_y=-\alpha(\bm U\odot\bm Y\oslash \bm {D})^T* \bm {e_n}+\beta \bm w
\end{split}
\end{equation}
The denominator matrix $\bm {D}$ is obtained by repeating the column vector $(1+\mbox{exp}{(\bm U\bm w+b \bm {e_n}))\odot \bm y}$ r times to form the matrix of size $n\times r$. $\bm Y$ is obtained with the same operator to vector $\bm y$.
\begin{equation}
\label{gradUy}
\begin{split}
\nabla_{u_{ij}}L_y=\alpha\frac{\mbox{exp}\big(-y_i(\sum_{j=1}^r w_ju_{ij}+b)\big)*(-1)*y_iw_j}{1+\mbox{exp}\big(-y_i(\sum_{j=1}^r w_ju_{ij}+b)\big)}
\end{split}
\end{equation}
It can be written in matrix form as follows.
\begin{equation}
\label{gradUymatrix}
\begin{split}
\nabla_{\bm U}L_y=-\alpha (\bm y {\bm w}^T)\oslash\bm {D}
\end{split}
\end{equation}
The denominator matrix $\bm D$ is the same with that in Eq.~\eqref{gradWmatrix}.
\begin{equation}
\label{gradb}
\begin{split}
&\nabla_{b}L_y=\alpha \sum_{i=1}^n\frac{\mbox{exp}\big(-y_i(\sum_{j=1}^r w_ju_{ij}+b)\big)*(-1)*y_i}{1+\mbox{exp}\big(-y_i(\sum_{j=1}^r w_ju_{ij}+b)\big)}+\beta b\\
&=-\alpha\sum_{i=1}^n\frac{y_i}{1+\mbox{exp}\big(y_i(\sum_{j=1}^r w_ju_{ij}+b)\big)}+\beta b.
\end{split}
\end{equation}
It can be written in matrix form as follows.
\begin{equation}
\label{gradbmatrix}
\begin{split}
\nabla_{b}L_y=-\alpha \bm {e_n}^T*\Big(y\oslash \big(1+\mbox{exp}(\bm U\bm w+b)\odot \bm y\big)\Big)+\beta b.
\end{split}
\end{equation}

\end{document}